\documentclass[5pt]{article}

\usepackage[letterpaper]{geometry}
\usepackage{spconf,amsmath,epsfig}
\usepackage{enumitem}
\usepackage{spconf,amsmath,epsfig}
\usepackage{epstopdf}
\usepackage{graphicx}
\usepackage{mathrsfs}
\usepackage[numbers,sort&compress]{natbib}
\usepackage{fancyhdr}
\begin{document}\sloppy

\def\x{{\mathbf x}}
\def\L{{\cal L}}

\title{Trajectory Factory: Tracklet Cleaving and Re-connection by\\ Deep Siamese Bi-GRU for Multiple Object Tracking}
%
\name{Cong Ma, Changshui Yang, Fan Yang, Yueqing Zhuang, Ziwei Zhang, Huizhu Jia, Xiaodong Xie\thanks{*The corresponding author, Changshui Yang is with the Institute of Digital Media, School of Electronic Engineering and Computer Science, Peking University.}}
\address{National Engineering Laboratory for Video Technology, Peking University, Beijing, China \\
\{Cong-Reeshard.Ma, csyang, fyang.eecs, zhuangyq, zhangziw, hzjia, donxie\}@pku.edu.cn}
%
%
%

\maketitle

\begin{abstract}
Multi-Object Tracking (MOT) is a challenging task in the complex scene such as surveillance and autonomous driving. In this paper, we propose a novel tracklet processing method to cleave and re-connect tracklets on crowd or long-term occlusion by Siamese Bi-Gated Recurrent Unit (GRU). The tracklet generation utilizes object features extracted by CNN and RNN to create the high-confidence tracklet candidates in sparse scenario. Due to mis-tracking in the generation process, the tracklets from different objects are split into several sub-tracklets by a bidirectional GRU. After that, a Siamese GRU based tracklet re-connection method is applied to link the sub-tracklets which belong to the same object to form a whole trajectory. In addition, we extract the tracklet images from existing MOT datasets and propose a novel dataset to train our networks. The proposed dataset contains more than 95160 pedestrian images. It has 793 different persons in it. On average, there are 120 images for each person with positions and sizes. Experimental results demonstrate the advantages of our model over the state-of-the-art methods on MOT16.
\end{abstract}
\vspace{-0.1cm}
\section{Introduction}
\vspace{-0.1cm}
\label{sec:intro}
Multi-object tracking (MOT) is a significant task of identifying each object and predicting their trajectories in a video sequence. It has a wide range of applications in computer vision, such as video surveillance, pedestrian flow analysis and autonomous driving. MOT based methods are aiming to address this problem by data association, which jointly optimize the matching process of bounding boxes detected by detector within the inter-frames of a sequence. One of the major applications of MOT focuses on pedestrian tracking.  The same individual has regular temporal or spatial cues in video. For example, a person has slight appearance, velocity and direction changes for monocular sequence in a single camera. Therefore, MOT usually depends on the combination of multiple cues (e.g. appearance, motion and interactions) to associate the similar bounding boxes. Although the performance is gradually improving at the MOT challenges~\cite{DBLP:journals/corr/MilanL0RS16}, the effectiveness of MOT is still limited by object detection quality, long-term occlusion and scene complexity. To solve this sophisticated problem, previous works aim to extract the competitive feature, design effective association metric and adopt reliable detector.
\begin{figure}
    \centering
    \includegraphics[width=8.5cm]{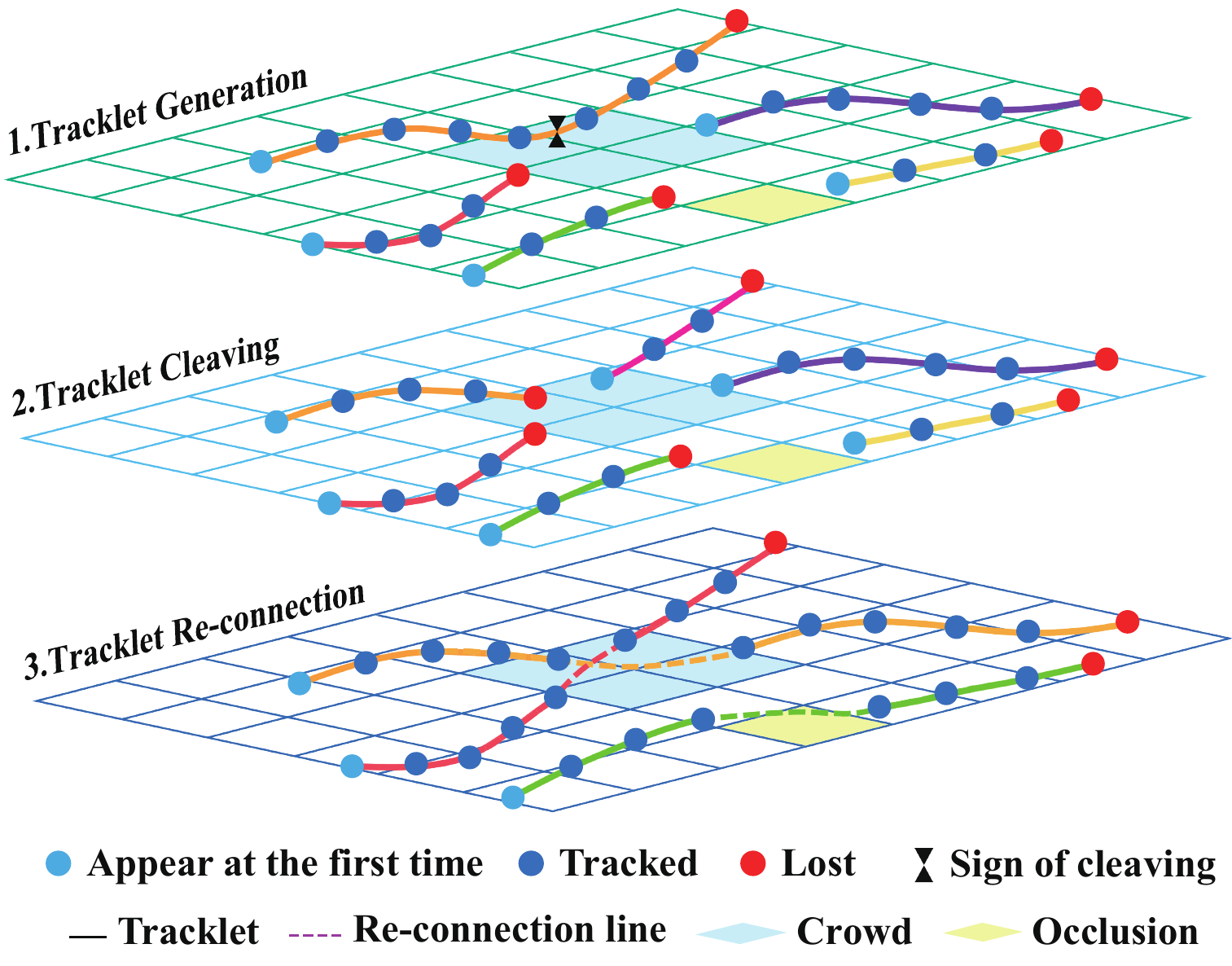}
    \caption{An example of our method: 1.Generating the tracklet candidates by appearance and motion model. 2.Cleaving the mistracked tracklets of a person. 3.Re-connecting the tracklets of the same person.}
    \vspace{-0.4cm}
\end{figure}

Tracking-by-detection is becoming dominant solutions for MOT , which compares feature and position of bounding boxes to link similar objects into trajectories. The aim of Tracking-by-detection is to search the optimal assignment from multiple cues within a set of bounding boxes. The person's appearance is a convincing cue for data association. Conventional algorithms tend to extract hand-crafted features. Currently, deep neural networks such as convolutional neural networks (CNN) and recurrent neural networks (RNN) have achieved state-of-the-art performance in MOT~\cite{tang2017multiple,chu2017online,sadeghian2017tracking}. CNN extracts the feature of bounding box to represent the pedestrian appearance. RNN is able to summarize the general characteristics of images from the same tracklet. These association methods are detection-to-detection or detection-to-tracklet. However, for a given tracklet with long-term occlusion, the detected pedestrian image may contain different degrees of occlusions. Thus, feature extraction on these occluded pedestrian images, even with subsequent complex matching techniques, is often not quite reliable.

In this paper, we propose a novel tracklet association to address the above problem by Siamese Bi-Gated Recurrent Unit(GRU), which is a tracklet-to-tracklet based method. GRU is the long-term version of RNN. Our method can be divided into three steps as illustrated in Fig.\ 1.

\begin{itemize}
  \item Tracklet Generation: We firstly utilize the non-maximum suppression (NMS) to eliminate redundant bounding boxes and associate them with less or none occlusion by appearance and motion cues to generate the high confidence tracklet candidates.
      \vspace{-0.1cm}
  \item Tracklet Cleaving: Due to occlusion, one tracklet may belongs to multiple persons. Therefore, we utilize bidirectional GRU to split the tracklet into several sub-tracklets and ensure that each sub-tracklet only belong to independent tracked person.
      \vspace{-0.1cm}
  \item Tracklet Re-connection: We extract features from each tracklet candidate or split sub-tracklet by siamese GRU. The tracklets are matched by using temporal and spatial cues and re-connected according to their similarity. At last, we fill the gap among matched tracklets by polynomial curve fitting to form the whole trajectory and smooth every trajectory by smoothing function.
\end{itemize}

\vspace{-0.4cm}
\section{Related Work}
\vspace{-0.2cm}
Multi-object tracking in videos has attracted great attention. The performance of MOT improves gradually at the MOT benchmark. Tracking-by-detection has become one of the most popular tracking frameworks. Among the methods of MOT,~\cite{tang2017multiple,xiang2015learning,choi2015near,kim2015multiple,chen2017enhancing} focus on designing an ingenious data association or multiple hypothesis.~\cite{schulter2017deep,Levinkov2017Joint,Maksai2016Globally} rely on network flow and graph optimization which are powerful approaches for tracking.~\cite{wang2016joint,bae2014robust} are presented to improve the tracklet association and tracklet confidence to achieve the tracklet task. The inter-relation of targets have multiple cues in a sequence including appearance, motion and interaction, which summarized are by~\cite{sadeghian2017tracking}. In addition,~\cite{le2016long,hong2016online} adopt the appearance model of some early traditional algorithms such as color histogram to represent the image feature, or~\cite{choi2015near,bae2014robust,yang2016temporal} utilize covariance matrix or hand-crafted keypoint features.~\cite{Henschel2017A} uses a novel multi-object tracking formulation to incorporate several detector into a tracking system.~\cite{kim2015multiple} extends the multiple hypothesis by enhancing detection model. The motion model expresses the rule of object movement, which are divided into linear position prediction~\cite{son2017multi} and non-linear position prediction~\cite{dicle2013way}. The interaction model describes the inter-relationship of different pedestrians in the same scene. ~\cite{hong2016online} designs the structural constraint by the location of people to optimize assignment. Recently, deep neural networks have been used gradually for tracking.~\cite{tang2017multiple,sadeghian2017tracking} train the CNN on the basis of person re-identification to extract the image features, and~\cite{son2017multi} utilizes the quadruplet loss to enhance the feature expression.~\cite{chu2017online} builds the CNN model to generate visibility maps to solve the occlusion problem. Following the success of RNN models for sequence prediction tasks,~\cite{alahi2016social} proposes social-LSTM to predict the position of each person in the scene.

\vspace{-0.1cm}
\section{Multi-Object Tracking Framework}
\vspace{-0.1cm}
Our solution aims at long-term occlusion and crowd which are difficult to track precisely. In this Section, the data association metric which generates tracklets from relative sparse scenario as the tracklet candidate is described in Section 3.1. We present how to estimate the tracklet reliability and split the unreliable tracklets in Section 3.2. Section 3.3 gives the traclets re-connection and association strategy, moreover, the training method of our network is also discussed.

\begin{figure}[h]
    \centering
    \includegraphics[width=8.5cm]{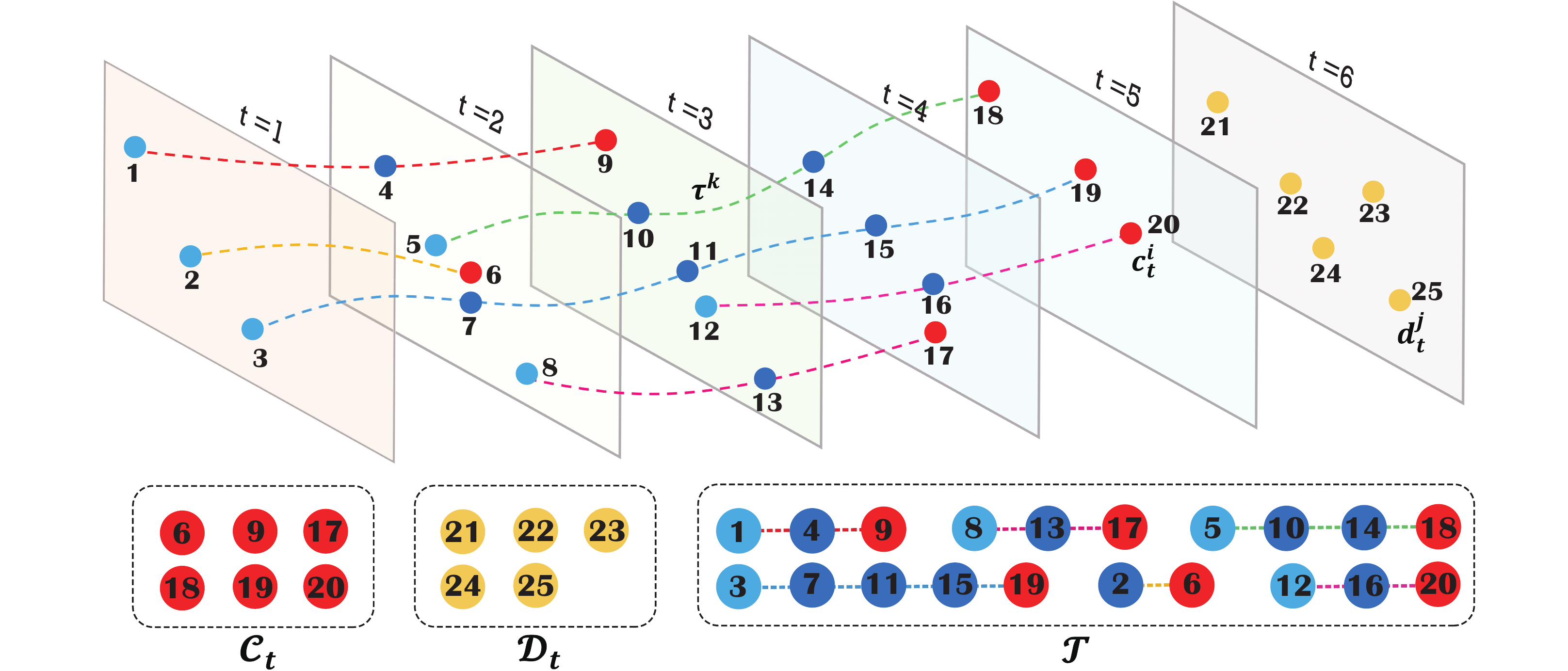}
    \vspace{-0.2cm}
    \caption{The structure of tracklets generation and the demonstration of the sets of $\mathcal{C}_{t}$, $\mathcal{D}_{t}$ and $\mathcal{T}$.}
\vspace{-0.5cm}
\end{figure}
\begin{figure*}
    \centering
    \includegraphics[width=17.5cm]{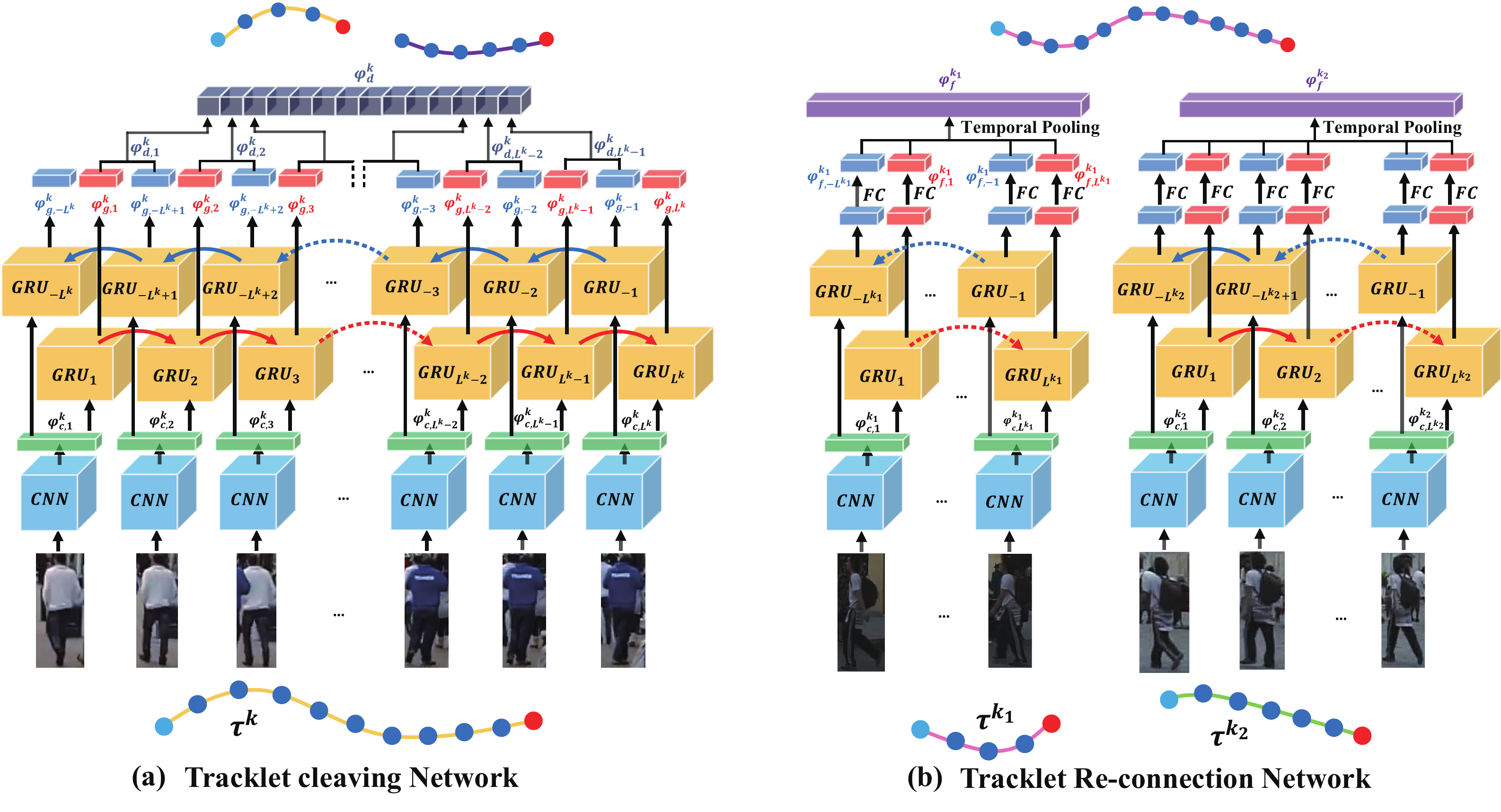}
    \vspace{-0.3cm}
    \caption{The architecture of tracklet cleaving and re-connection network, (a) Cleaving the tracklets by bidirectional outputs of GRU, (b) Re-connecting the tracklets by the features of siamese GRU.}
    \label{Fig13}
\vspace{-0.5cm}
\end{figure*}
\subsection{Tracklet Generation}
\vspace{-0.1cm}
Firstly, we execute a simple multi-object tracking algorithm to generate tracklets. We choose the target which is easy to track in order to produce the high-confidence tracklets. So we denote the set of detection bounding boxes $\mathcal{D}_t$ ($d_t^k \in \mathcal{D}_t $)and the set of tracked objects candidates $\mathcal{C}_{t}$ ($c_n^k \in \mathcal{C}_{t} ; n \leq t,\mathcal{C}_{t}=\mathcal{C}_{t-1}\bigcup\mathcal{D}_{t-1}$), where $d_t^k$ and $c_t^k$ are $k$-th detection and candidate in frame $t$, respectively. To connect the candidate and detection within inter-frames, we match the candidates ${c_t^k}$ and ${d_t^k}$ in a bipartite graph with Hungarian algorithm~\cite{Sahbani2017Kalman}. The bipartite graph $G=(\mathcal{V},\mathcal{E})$ whose node $V$ are divided into left part $\mathcal{C}_t\in\mathcal{V}_L$ and right part $\mathcal{D}_t\in\mathcal{V}_R$, $e_{ij}\in\mathcal{E}$ is the edge of $c_t^i$ and $d_t^j$. The tracked objects are defined as 7 dimensions $[t, id, x, y, w, h, s]$ that contain the tracklet id by tracker, the object time, the center position $(x,y)$, width and height of the bounding box, and the state of the tracklet. The state of tracklet includes "tracked", "lost" and "quitted", which are similar to Markov Decision Processes~\cite{xiang2015learning} (as described in Fig.\ 2). The detail of state transition is introduced in Section 3.3. And then we obtain the set of tracklets $\mathcal{T}$ ($\tau^k \in \mathcal{T} $) in the whole sequence. The formulation of optimized graph is given by
\begin{eqnarray}
argmin\sum_{e_{ij}\in\mathcal{E}}\mathcal{S}(c_t^i,d_t^i)e_{ij}
\vspace{-0.2cm}
\end{eqnarray}
where $\mathcal{S}(c_t^i,d_t^i)$ indicates the cost function with $c_t^i$ and $d_t^j$. In addition, $e_{ij}$ is the indicator parameter $e_{ij}\in \{0,1\}$.The cost function is defined as
\begin{eqnarray}
\mathcal{S}(c_t^i,d_t^i)=\alpha F_a(c_t^i,d_t^i)+\beta F_m(c_t^i,d_t^i)
\vspace{-0.6cm}
\end{eqnarray}
\begin{eqnarray}
F_a(c_t^i,d_t^i)=\parallel f_{c_t^i}-f_{d_t^i}\parallel_2^2,  F_m(c_t^i,d_t^i)=\parallel \hat{p}_{c_t^i}-p_{d_t^i}\parallel_2^2
\end{eqnarray}
where $F_a(c_t^i,d_t^i) $ denotes the appearance cue which calculates the Euclidean distance and $L_2$ normalized to measure the similarity of $c_t^i$ and $d_t^j$. Furthermore, the appearance features $f_{c_t^i}, f_{d_t^i}$ of a person are created by Convolutional Neural Network (CNN). $\alpha, \beta$ are the weight coefficients of the function.  $F_m(c_t^i,d_t^i) $ indicates the motion cue, and the function compares the distance between the detection position $p_{d_t^i}$ and candidate prediction position $\hat{p}_{c_t^i}$, which is defined in 4 dimensions $[\hat{x},\hat{y},\hat{w},\hat{h}]$ that stand for the prediction of $x,y$-coordinate, weight and height, respectively. The prediction position by output of Long Short-Term Memory (LSTM) depends on historical position as the input of LSTM. Some more details of CNN and LSTM are discussed in Section 4.1.

\vspace{-0.2cm}
\subsection{Tracklet Cleaving}
\vspace{-0.1cm}
After tracklet generation, we have the coarse set of tracklet  $\mathcal{T}$ in sequence. However, the tracker in Section 3.1 may mis-track the wrong person when two persons cross each other. To guarantee the tracklet with the single person, we design a bidirectional output Gated Recurrent Unit (GRU) to estimate the tracklet reliability and cleave the false tracklets in time. The reliable tracklets $\mathcal{T}^+$ and unreliable tracklets $\mathcal{T}^-$ are defined as
\begin{eqnarray}
\left\{\begin{matrix}
\tau^k\in \mathcal{T}^+ &\forall i,j, &r_i^k,r_j^k\in \tau^k , r_i^k (id)\equiv r_j^k (id)\\
\tau^k\in \mathcal{T}^- &\exists i,j, & r_i^k,r_j^k \in \tau^k , r_i^k (id)\neq r_j^k (id)
\end{matrix}\right.
\end{eqnarray}
where $ r_i^k,r_j^k$ are the $i$-th, $j$-th element on tracklet $ \tau^k$. All of the tracklets $\tau^k \in \mathcal{T}$ are fed into the bi-GRU to distinguish whether the tracklet is reliable, or find out the split position of the unreliable tracklet. The tracklet cleaving network (bidirectional-GRU) is shown in Fig.\ 3. First of all, we utilize the CNN to extract the image features $\varphi_{c,i}^k,\ i\in[1,L^k]$ from the tracklet. Secondly, all the features  $\varphi_{c,i}^k$ is inputted the forward-GRU and backward-GRU respectively. Both GRUs have the shared weights, and the output is $\varphi_{g,i}^k,\ i\in[-L^k,-1]\cup[1,L^k]$, the positive and negative values stand for forward and backward feature from GRU. And then, we calculate the adjacent vector distance between the forward and the backward (e.g. length=10, $\{\varphi_{g,1}^k,\varphi_{g,-9}^k\},\{\varphi_{g,2}^k,\varphi_{g,-8}^k\}, ...$ ) as a series of feature distance to combine a 1 $\times (L^k-1)$ vector $\varphi_d^k$:
\begin{eqnarray}
\begin{split}
&\varphi_{d,i}^k=\parallel\varphi_{g,i}^k-\varphi_{g,i-L^k}^k\parallel_2^2,\ i\in[1,L^k-1]\\
&\quad\quad \varphi_d^k=([\varphi_{d,1}^k,\varphi_{d,2}^k,...,\varphi_{d,L^k}^k])\\
\end{split}
\vspace{-0.2cm}
\end{eqnarray}

The algorithm calculates the distance $\varphi_{d,i}^k$ between the features from the left and the right to current position and search the maximum disparity from these distances. The final output of the cleaving network is a single vector $\varphi_d^k$, which can find the most suitable splitting point by the position of peak value. However, if all of the vector values are less than the threshold, the tracklet includes the same person. The example is described in Fig.\ 4. In this figure, the input is a unreliable tracklet (the length is 10) which includes the white coat person at the front part and the blue coat person at the latter part. The network calculates every adjacent feature distance $\varphi_{d,i}^k, i\in[1,9]$ from left and right, and find the maximum distance to define the best splitting point($\varphi_{d,4}^k$). So our cleaving network not only distinguishes the tracklet availability, but also cleaves the unreliable tracklet.

GRU is a long-term version of RNN. The advantage of RNN is able to summarize the general characteristics with the same person and eliminate occlusion in order to obtain preferable feature expression. The pre-train model of cleaving network is the half of the re-connection network (shown in Fig.\ 3 (b)). The details of training strategy is described in Section 3.3.
\begin{figure}
    \centering
    \includegraphics[width=8.5cm]{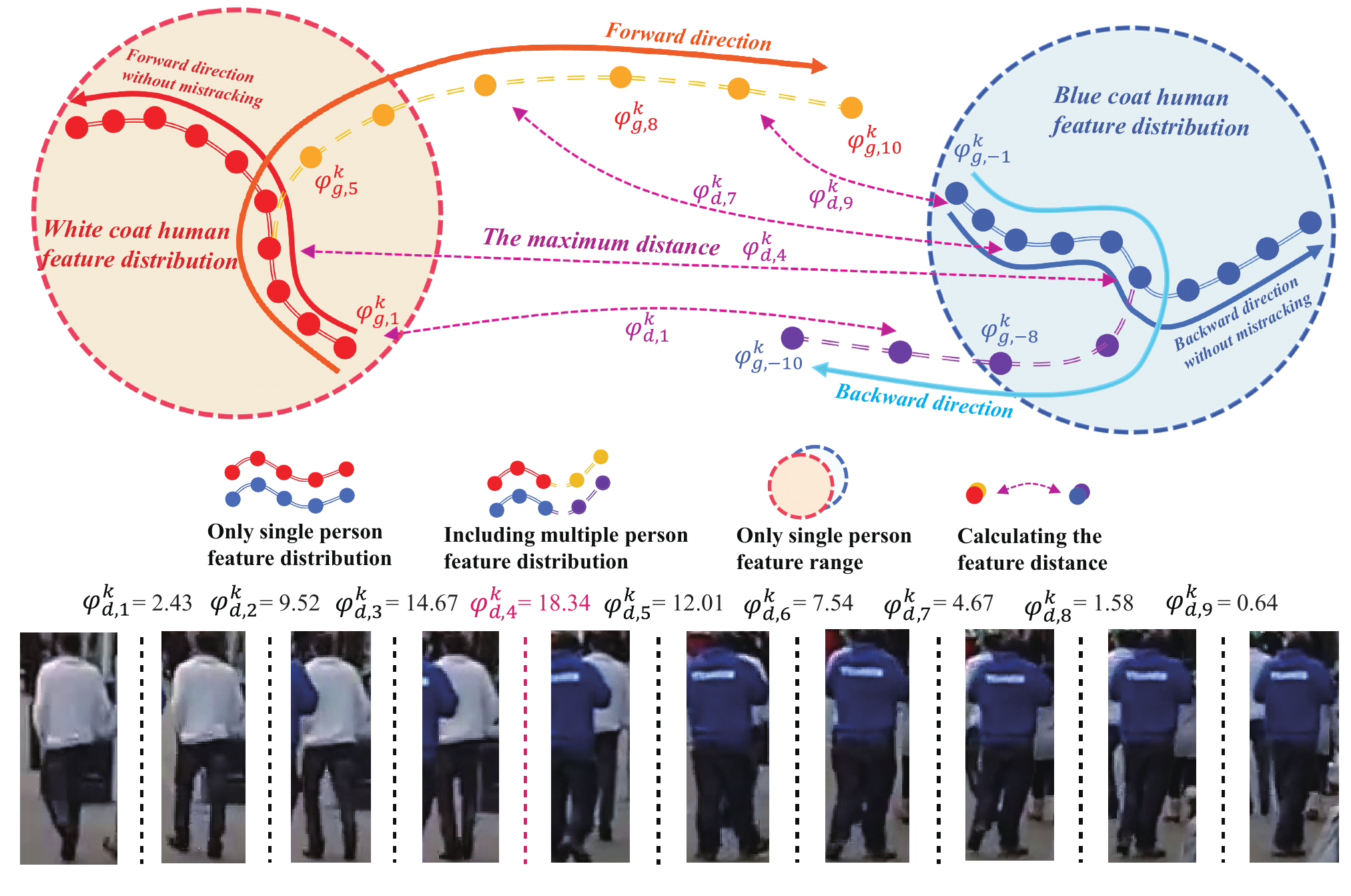}
    \vspace{-0.3cm}
    \caption{The explanation of the cleaving network: Top of the dots indicate feature distribution for two persons, and the arrows denote normal tendency with a single person and disrupted tendency by each other. Bottom of the figure is the unreliable tracklet corresponding to the top figure.}
\vspace{-0.6cm}
\end{figure}
\vspace{-0.4cm}
\subsection{Tracklet Re-connection}
\vspace{-0.2cm}
We construct the Deep Siamese Bi-GRU to perform the cleaving and re-connection tasks. To obtain the competitive feature descriptor, we combine various losses to reduce the within-class distance and enlarge the between-class distance simultaneously. Our network is designed with the verification loss and identification loss at each GRU output. The loss is defined as:
\vspace{-0.1cm}
{\setlength{\abovedisplayskip}{2mm}
\begin{eqnarray}
\vspace{-0.1cm}
&\mathcal{L}_{sum}=\mathcal{L}_{glo}+\mathcal{L}_{loc}
\vspace{-0.2cm}
\end{eqnarray}}
\ \ \ Where the $\mathcal{L}_{glo}$ and $\mathcal{L}_{loc}$ indicate the global loss and local loss of the network, respectively. We use the contrastive loss by Euclidean distance for the verification and the cross-entropy losses in the multi-classification task for the identification. The details of the losses are shown as:
\begin{eqnarray}
\begin{split}
&E(\varphi_f^{k_1},\varphi_f^{k_2})=y\parallel\varphi_f^{k_1}-\varphi_f^{k_2}\parallel_2^2\\
&\qquad\qquad+(1-y)max\{0,(\eta-\parallel\varphi_f^{k_1}-\varphi_f^{k_2}\parallel_2^2)\}\\
\end{split}
\end{eqnarray}
{\setlength{\abovedisplayskip}{-2mm}
\begin{eqnarray}
\begin{split}
F(\varphi_f^k)=\sum_{i=1}^{K}-p_ilog(\hat{p_i}),\quad \hat{p_i}=softmax(\varphi_f^k)
\end{split}
\end{eqnarray}}
\ \ \ where $\varphi_f^k$ indicates the output feature of GRU $\varphi_g^k$ after fully-connected (FC) layer and ReLU. $E(\varphi_i,\varphi_j)$ is the contrastive function, $y\in\{0,1\}$ is the label indicator, $\eta$ is a margin constant. $F(\varphi)$ denotes the multi-classification cross-entropy function. The representation of loss can be formulated as follows:
\begin{eqnarray}
\begin{split}
&\ \mathcal{L}_{glo}=\lambda_v\mathcal{L}_{v}+\lambda_{id}(\mathcal{L}_{id1}+\mathcal{L}_{id2})\\
&\qquad=\lambda_v E(\varphi_f^{k_1},\varphi_f^{k_2})+\lambda_{id}(F(\varphi_f^{k_1})+F(\varphi_f^{k_2}))
\end{split}
\end{eqnarray}
{\setlength{\abovedisplayskip}{-1mm}
\begin{eqnarray}
\varphi_f^k=\frac{1}{2L^k}(\sum_{i=-L^k}^{-1}\varphi_{f,i}^k+\sum_{j=1}^{L^k}\varphi_{f,j}^k)
\end{eqnarray}}
where $\varphi_f^k$ is the temporal pooling~\cite{mclaughlin2016recurrent} of each output of GRU.
\begin{eqnarray}
&\mathcal{L}_{loc}=\lambda_{loc\_v}\mathcal{L}_{loc\_v}+\lambda_{loc\_id}\mathcal{L}_{loc\_id}
\end{eqnarray}
\begin{eqnarray}
\begin{split}
&\mathcal{L}_{loc\_v}=\parallel\varphi_{f,1}^{k_1}-\varphi_{f,L^{k_1}}^{k_1}\parallel_2^2+\parallel\varphi_{f,1}^{k_2}-\varphi_{f,L^{k_2}}^{k_2}\parallel_2^2\\
&\qquad\ \ -\parallel\varphi_{f,1}^{k_2}-\varphi_{f,1}^{k_2}\parallel_2^2-\parallel\varphi_{f,L^{k_1}}^{k_1}-\varphi_{f,L^{k_2}}^{k_2}\parallel_2^2+\delta
\end{split}
\vspace{-0.2cm}
\end{eqnarray}
\begin{eqnarray}
\begin{split}
\vspace{-0.1cm}
\mathcal{L}_{loc\_id}=\sum_{k\in{k_1,k_2}}(\sum_{i=-L^k}^{-1}F(\varphi_{f,i}^k)+\sum_{j=1}^{L^k}F(\varphi_{f,j}^k))
\end{split}
\end{eqnarray}

\vspace{-0.1cm}
$\lambda$ is the loss weight coefficient. $\mathcal{L}_{v}$, $\mathcal{L}_{id*}$, $\mathcal{L}_{loc\_v}$ and $\mathcal{L}_{loc\_id}$ denote the verification and identification loss of global and local respectively. $\mathcal{L}_{loc\_v}$ is similar to triplet loss (refer to~\cite{son2017multi}), including the disparity of head and tail of the tracklet, head between different tracklets and tail between different tracklets. $\delta$ is the threshold of margin. $\mathcal{L}_{loc\_id}$ is the multi-classification task for each output.

After training the re-connection network, we can cleave or match the tracklets. For cleaving the tracklet, we calculate the peak value of the feature which is concatenated by $\varphi_{d,i}^k$, $\varphi_d^k=([\varphi_{d,1}^k,\varphi_{d,2}^k,...,\varphi_{d,L^k}^k])$. For re-connecting the tracklet, we match the temporal pooling features $\varphi_f^k$ to compare the distance between the tracklets.
\vspace{0.1cm}

\noindent\textbf {Tracklet Association}. This process is the tracklet assignment from the set of tracklet $\mathcal{T}^+$ after cleaving network. The matching of tracklets $\tau^k\in \mathcal{T}^+$ are restricted previously by temporal and spatial constraints. The constraint of matching is shown as:
\begin{table*}[t]\footnotesize
\vspace{-0.2cm}
\begin{center}
\caption{Results on the MOT16 test dataset (G: Generation C: Cleaving R: Re-connection A: Association)} \label{tab:cap}
\vspace{0.2cm}
\begin{tabular}{c|c|c|c|c|c|c|c|c|c}

  Tracker & MOTA$\uparrow$ & IDF1$\uparrow$ & MT$\uparrow$ & ML$\downarrow$ & FP$\downarrow$ & FN$\downarrow$ & IDSw.$\downarrow$ & Frag$\downarrow$ & Hz$\uparrow$
  \\
  \hline
  \hline
  QuadMOT16~\cite{son2017multi} & 44.1  & 38.3 & 14.6\% & 44.9\% & 6388 & 94775 & 745 & 1096 & 1.8 \\
  EDMT~\cite{kim2015multiple} & 45.3  & 47.9 & 17.0\% & 39.9\% & 11122 & 87890 & 639 & 946 & 1.8 \\
  MHT\_DAM~\cite{Henschel2017A} & 45.8  & 46.1 & 16.2\% & 43.2\% & 6412 & 91758 & 590 & 781 & 0.8 \\
  STAM16~\cite{chu2017online} & 46.0  & 50.0 & 14.6\% & 43.6\% & 6895 & 91117 & 473 & 1422 & 0.2 \\
  NOMT~\cite{Henschel2017A} & 46.4  & \textbf{53.3} & 18.3\% & 41.4\% & 9753 & 87565 & \textbf{359} & 504 & 2.6 \\
  AMIR~\cite{sadeghian2017tracking} & 47.2  & 46.3 & 14.0\% & 41.6\% & \textbf{2681} & 92856 & 774 & 1675 & 1.0 \\
  NLLMPa~\cite{wang2016joint} & 47.6  & 50.9 & 15.2\% & 38.3\% & 9253 & \textbf{85431} & 792 & 1858 & 18.5 \\
  FWT~\cite{Henschel2017A} & 47.8  & 44.3 & \textbf{19.1}\% & \textbf{38.2}\% & 8886 & 85487 & 852 & 1534 & 0.6 \\
  LMP~\cite{tang2017multiple} & \textbf{48.8}  & 51.3 & 18.2\% & 40.1\% & 6654 & 86245 & 481 & \textbf{595} & 0.5 \\
  \hline
  \hline
  GCRA\_G(Ours) & 47.4  & 41.4 & 14.4\% & 39.8\% & 7516 & 87219 & 1147 & 1156 & 7.5 \\
  GCRA\_G+C+R+A(Ours) & 48.2  & 48.6 & 12.9\% & 41.1\% & 5104 & 88586 & 821 & 1117 & 2.8\\
  \hline
\end{tabular}
\end{center}
\vspace{-0.7cm}
\end{table*}
\vspace{-0.1cm}
\begin{eqnarray}
&IOU(\hat{r}_{L^i+\Delta t_{i,j}}^i,r_1^j)=\frac{area(B(\hat{r}_{L^i+\Delta t_{i,j}}^i)\cap B(r_1^j))}{area(B(\hat{r}_{L^i+\Delta t_{i,j}}^i)\cup B(r_1^j))}>0\ \ \
\end{eqnarray}
where
\vspace{-0.3cm}
\begin{eqnarray}
\begin{split}
&\quad\quad\ \ \hat{r}_{L^i+\Delta t_{i,j}}^i(x,y)= (\bar{\tau^i}(v)+\mu)\cdot \Delta t_{i,j} \\
&\quad\quad \ \Delta t_{i,j}= (r_1^j(t)-r_{L^i}^i(t)) \\
&\qquad s.t. \ \ r_1^j(t)> r_{L^i}^i(t),\ \tau^i(s),\tau^j(s)\neq quitted
\end{split}
\vspace{-0.3cm}
\end{eqnarray}

\vspace{-0.1cm}
$B(\cdot)$ is the bounding box of object $[x,y,w,h]$. $\bar{\tau^i}(v)$ is the average velocity of tracklet $\tau^i$, $\mu$ is velocity constant. $r_{L^i}^i$ and $r_1^j$ indicate the tail of tracklet $\tau^i$ and the head of tracklet $\tau^j$ respectively. The states of tracklet node  $r_i^k(s)$ include "tracked" "lost" and "quitted". The node is unmatched more than one frame, which is transferred to "lost" until the node is matched by next frame node and transfer to "tracked".The node is quitted when the node position will be out of the boundary. In addition, tracklets also has states $\tau^k(s)$, which depend on the last node state of this tracklet. For tracklet association, we give up the quitted tracklets, and re-connection network associates the tracklets which meet the temporal-spatial conditions. Lastly, we re-connect the tracklets which satisfy with the constraint by the output $\varphi_f^k$.
\vspace{-0.2cm}
\section{Experimental Results}
\vspace{-0.1cm}
\subsection{Implementation Details}
\vspace{-0.1cm}
In our experiments, our networks consist of CNN, LSTM and GRU. For tracklet generation, we train the Siamese-CNN network of appearance model with ResNet-50, the images are resize to 224 $\times$ 224 from the Re-identification dataset Market1501~\cite{zheng2015scalable} and the output of CNN produced a 1024-dimensional vector to describe the image. In addition , the inputs of
the LSTM network for motion model is a series of 4-dimensional vector $[x, y, w, h]$ with a tracklet (the length of input vectors $L\in[3,10]$), and the LSTM output is the prediction of the position and size $[\hat{x},\hat{y}, \hat{w}, \hat{h}]$.
For tracklet cleaving and re-connection, the model is a deep siamese bi-GRU, which includes four hidden-layers and the maximum length of GRU is 120 frames. The inputs of the GRU is a series of 128-dimensional features by CNN which
is a wide residual networks (WRN)~\cite{zagoruyko2016wide}. The outputs of the GRU $\varphi_{g,i}^k, i\in[-L^k,-1]\cup[1,L^k]$ are also 128-dimensional vectors, which are fed to FC network for verification loss and for comparison of the corresponding features for
classification loss. We use the AdamOptimizer~\cite{kingma2014adam} to train our network. The training dataset is extracted from dataset PathTrack~\cite{manen2017pathtrack} and video re-identification dataset MARS~\cite{zheng2016mars}.
\vspace{-0.1cm}
 \begin{table}[h]\scriptsize
\vspace{-0.4cm}
\begin{center}
\caption{MOTA of each MOT16 sequences} \label{tab:cap}
\vspace{-0.1cm}
\begin{tabular}{c|c|c|c|c|c|c|c}

   \hline
  Sequence  & 01& 03 & 06 & 07 & 08 & 12 & 14 \\
    \hline
  Static(s)\&Moving(m)  & s& s & m & m & s & m & m \\
  \hline
  \hline
  QuadMOT16~\cite{son2017multi} & 30.8  & 51.0 & 49.2& 41.9 & 29.9 & 38.0 & 24.0  \\
  EDMT~\cite{kim2015multiple} & 35.3  & 51.2 & 49.8 & \textbf{46.1} & 32.3 & 43.1 & 24.9  \\
  MHT\_DAM~\cite{Henschel2017A} & 35.8  & 52.7 & 49.1 & 39.3 & 33.2 & 44.3 &26.1  \\
  STAM16~\cite{chu2017online} & 35.7&53.8&48.4&38.0&32.3&42.3&24.6  \\
  NOMT~\cite{Henschel2017A} & 34.2&53.0&51.3&44.9&\textbf{36.7}&39.3&23.5 \\
  AMIR~\cite{sadeghian2017tracking} & 37.8&53.8&49.2&45.5&32.5&40.4&\textbf{29.4} \\
  NLLMPa~\cite{wang2016joint} & 30.7&56.4&49.8&40.7&33.3&43.3&23.3  \\
  FWT~\cite{Henschel2017A} & 33.6&55.7&51.8&40.3&35.1&44.67&24.7  \\
  LMP~\cite{tang2017multiple} &39.9&56.1&\textbf{52.3}&43.1&33.8&43.7&28.8 \\
  \hline
  \hline
    Rank  & \textbf{1} & \textbf{1} & 10 & 3 & 2 & 8& 3  \\
  GCRA(Ours)  & \textbf{42.5} & \textbf{56.7} & 35.9 & 44.1 & 35.4 & 39.6& 28.3  \\

  \hline
\end{tabular}
\end{center}
\vspace{-0.9cm}
\end{table}

\vspace{-0.3cm}
\subsection{Results of Multi-Object Tracking}
\vspace{-0.1cm}
\textbf {Evaluation Metrics}.\ The MOTChallenge Benchmark depends on multiple evaluation index of trackers. These metrics~\cite{bernardin2008evaluating}~\cite{ristani2016performance} include Multiple Object Tracking Accuracy (MOTA), ID F1 Score (IDF1), Mostly tracked targets (MT), Mostly lost targets (ML), False Positives (FP), False Negatives (FN), Identity Switches (IDSw.), the total number of Fragment (Frag) and Processing Speed (Hz).

\noindent\textbf {MOTChallenge Benchmark}. We evaluated performance of our method on MOT16~\cite{DBLP:journals/corr/MilanL0RS16}. The sequences of dataset are captured from surveillance, hand-held shooting and driving recorder by static camera and moving camera.

\noindent\textbf {Result Comparison}. We compare the state-of-the-art methods on MOT16. The result of MOT benchmark is presented in Table 1. GCRA\_G is the tracklet generation in this paper, and the GCRA is our final method. Obviously, our method achieves higher performance of MOTA which is the primary evaluation metric. The result of static camera sequence is better than others especially, but moving camera is unsatisfactory because the temporal and spatial constraints are not suitable for it(as shown in Table 2).
\vspace{-0.3cm}
\begin{figure*}[t]
    \centering
    \includegraphics[width=16cm]{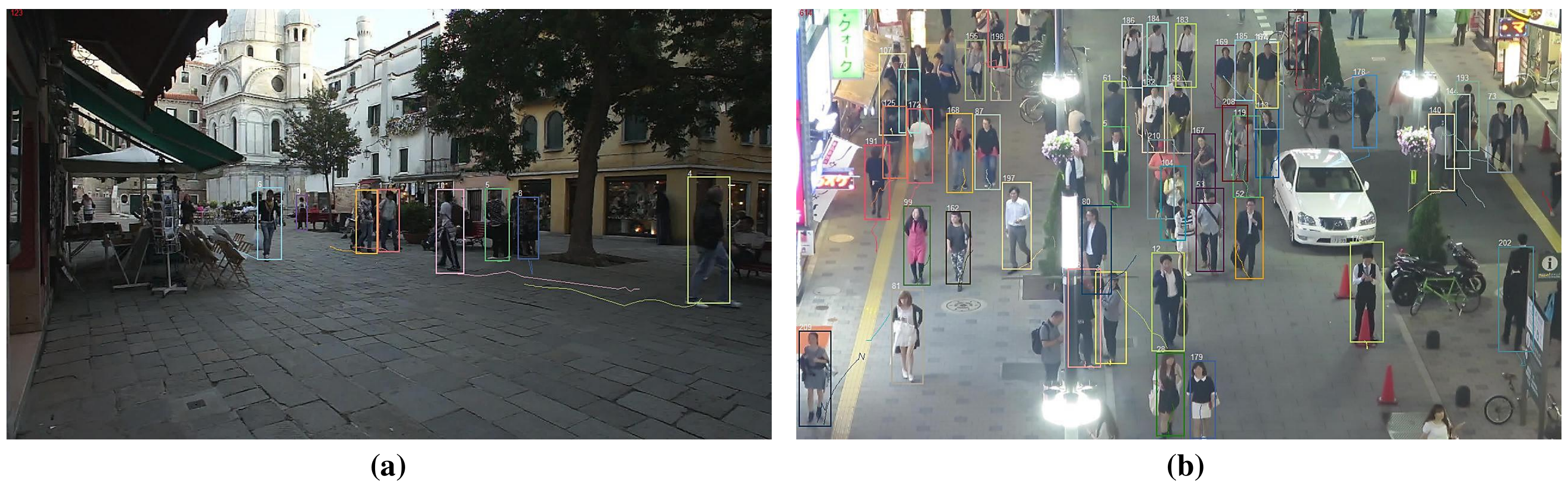}
    \vspace{-0.5cm}
    \caption{Qualitative results on the MOT16 benchmark. (a): MOT16-01 (b): MOT16-03}
\vspace{-0.3cm}
\end{figure*}
\section{Conclusion}
\vspace{-0.1cm}
We propose a novel tracklet association scheme to cleave and re-connect the tracklets on crowd or long-term occlusion by Deep Siamese Bi-GRU. The method calculates each output of bidirectional GRU to search the suitable split position and match the tracklets to reconnect the same person. For training, we extracted the tracklet dataset from existing MOT datasets for training our frameworks. Our proposal has better performance for static camera such as surveillance. The algorithm achieves 48.2\% in MOTA that approaches the state-of-the-art methods on MOT16 benchmark dataset. The qualitative result is shown in Fig.\ 5.
\vspace{-0.3cm}
\section{Acknowledgement}
\vspace{-0.2cm}
This work is partially supported by the National Science Foundation of China No.(61421062, 61602011), the Major National Scientific Instrument and Equipment Development Project of China under contract No.2013YQ030967, National Key Research and Development Program of China (2016YFB0401904) and NVIDIA NVAIL program.
\vspace{-0.2cm}

\vspace{-0.2cm}
\section{References}
\renewcommand\refname{}
\vspace{-10.5mm}
\bibliographystyle{IEEEbib}
\bibliography{icme2018template}
\vspace{-0.2cm}
\end{document}